# INTEGRATING SEMI-SUPERVISED LABEL PROPAGATION AND RANDOM FORESTS FOR MULTI-ATLAS BASED HIPPOCAMPUS SEGMENTATION

*Qiang Zheng[1, 2, 3], Yong Fan[1]*
*and for the Alzheimer's Disease Neuroimaging Initiative[*]*

[1]Department of Radiology, Perelman School of Medicine, University of Pennsylvania, Philadelphia, PA, 19104, USA
[2]School of Computer and Control Engineering, Yantai University, Yantai, 264005, China
[3]National Laboratory of Pattern Recognition, Institute of Automation, Chinese Academy of Sciences, Beijing, 100190, China

**ABSTRACT**

A novel multi-atlas based image segmentation method is proposed by integrating a semi-supervised label propagation method and a supervised random forests method in a pattern recognition based label fusion framework. The semi-supervised label propagation method takes into consideration local and global image appearance of images to be segmented and segments the images by propagating reliable segmentation results obtained by the supervised random forests method. Particularly, the random forests method is used to train a regression model based on image patches of atlas images for each voxel of the images to be segmented. The regression model is used to obtain reliable segmentation results to guide the label propagation for the segmentation. The proposed method has been compared with state-of-the-art multi-atlas based image segmentation methods for segmenting the hippocampus in MR images. The experiment results have demonstrated that our method obtained superior segmentation performance.

## 1. INTRODUCTION

Accurate segmentation of the hippocampus in magnetic resonance (MR) images is often needed in neuroimaging studies of neuropsychiatric disorders. Since manual delineation of the hippocampus is time-consuming, it is desirable to automatically segment the hippocampus with high accuracy, especially in large scale studies.

Among existing automatic medical image segmentation methods, multi-atlas based image segmentation (MAIS) has been widely adopted for the hippocampus segmentation [1]. Particularly, the MAIS methods align atlas images and their segmentation labels to a target image to be segmented, and then the aligned segmentation labels are fused to obtain a segmentation result for the target image.

In the past years, a variety of MAIS methods have been developed to improve the image segmentation [1]. In addition to improving image registration [2-6], most MAIS methods are developed to improve the label fusion [7-14]. In particular, majority voting (MV) might be the simplest label fusion method [15], and sophisticated label fusion strategies have been built upon a nonlocal patch-based label fusion strategy [16], such as joint label fusion [9], multi-scale and multi-feature label fusion based on optimized image patch matching [12], dictionary learning [10], and metric learning [7]. The label fusion could also be modeled as a pattern recognition problem [11, 14]. In the pattern recognition setting, image patches of the aligned atlases are used as training data to build prediction models using support vector machines (SVM) [11], artificial neural networks (ANNs) [17], or linear regression model [8], and then the trained models are used to predict segmentation labels of image patches of the target image.

Regardless differences of the existing label fusion methods, they typically implement the label fusion of different voxels independently without taking into consideration correlations among voxels of the target image, and therefore their performance might be degraded by imaging noise. Furthermore, the label fusion might also be hampered by image discrepancy across subjects that cannot be fully handled by the image registration, particularly differences between the target image and the atlas images [18, 19]. Although these problems have been addressed in different methods for medical image segmentation problems [18-20], a unified solution is desirable.

To improve the hippocampus segmentation, we integrate a semi-supervised label propagation (SSLP) method and a supervised random forests (RF) method in the MAIS framework [21, 22]. Particularly, a probabilistic image segmentation of each voxel in the target image is obtained using a local prediction model built upon its neighboring image patches of the atlas images using RF regression in the MAIS framework. Then, a semi-supervised label propagation method is adopted to segment the target image by propagating reliable segmentation information within the target image regularized by local and global image consistency [18, 19, 21]. The segmentation information propagates within the target image itself with an information-balance-weighting scheme, rather than propagating information among the target image and the atlas images [20], which may suffer from the inter-subject image inconsistency. The method has been validated for the hippocampus segmentation based on imaging data and segmentation labels provided by the EADC–ADNI (European Alzheimer's Disease Consortium and Alzheimer's Disease Neuroimaging Initiative) harmonized segmentation protocol (www.hippocampal-protocol.net) [23].

---
[*] Data used in preparation of this article were obtained from the Alzheimer's Disease Neuroimaging Initiative (ADNI) database (adni.loni.usc.edu). As such, the investigators within the ADNI contributed to the design and implementation of ADNI and/or provided data but did not participate in analysis or writing of this report. A complete listing of ADNI investigators can be found at: http://adni.loni.usc.edu/wp-content/uploads/how_to_apply/ADNI_Acknowledgement_List.pdf

## 2. METHODS

Our method consists of an MAIS method for generating a probabilistic segmentation result using a RF regression method and a SSLP method for computing reliable image segmentation for the hippocampus. The method is referred to as RF-SSLP.

### 2.1 MAIS based on random forest regression

Given a target image $I$ and $N$ registered atlas images and their labels $A_i = (I_i, L_i), i = 1, \cdots, N$, where $I_i$ and $L_i$ are the $i^{th}$ atlas and label images respectively, RF regression models are built to generate a probabilistic segmentation result of the target image in a pattern recognition framework [11]. In order to label each voxel in the target image, image patches in its neighborhood $N(x)$ with $(2r + 1) \times (2r + 1) \times (2r + 1)$ voxels are extracted from each atlas image as training samples. From each of the image patches, texture image features are computed using the method described in [11]. We obtain $(2r + 1)^3 \times N$ training samples $\{(\vec{f}_{i,j}, l_{i,j}) | i = 1, \cdots, N, j \in N(x)\}$, where $\vec{f}_{i,j}$ is a feature vector with label $l_{i,j} \in \{+1, -1\}$. Finally, a RF regression model is trained for predicting segmentation label of the voxel under consideration. To build the regression model on balanced training samples, we select the same number ($k$) positive and negative samples, most similar to the image patch to be segmented, to train the regression model [11].

The RF regression model is an ensemble of regression trees that are built upon the training data using randomly sampling, with 2 parameters: $N_{Tree}$ (the number of trees) and $N_{Split}$ (the number of predictors sampled for splitting at each node) [22]. Once $N_{Tree}$ regression trees are trained, they are applied to image features $f_x$ of voxel $x$ to predict its segmentation label:

$$P(f_x) = \frac{1}{N_{Tree}} \sum_{i=1}^{N_{Tree}} T_i(f_x) \quad (1)$$

where $T_i$ is the $i^{th}$ regression tree, and $P$ is the prediction of segmentation label. In this study, we treat the output of a regression model as a "probabilistic" segmentation label.

A specific regression model is built for each voxel of the target image, similar to the local label learning (LLL) method [11], and a probabilistic segmentation map is finally obtained for the target image. The RF based label fusion method is referred to as LLL-RF-regression hereafter. A binary segmentation results can be obtained by thresholding the probabilistic map. However, probabilistic segmentation results with smaller absolute label values might be unreliable. Furthermore, the segmentation results might be degraded by imaging noise since they are obtained for different voxels separately. The segmentation result can be improved by the semi-supervised label propagation method [18, 19, 21].

### 2.2 Semi-supervised label propagation for the hippocampus segmentation

A graph theory based label propagation method is adopted to improve the image segmentation [21]. Particularly, the label propagation will update the probabilistic segmentation map, regularized by local and global image consistency, by minimizing

$$E(L) = L^T(I - S)L + \alpha(L - L_0)^T(L - L_0), \quad (2)$$

where $L_0$ and $L$ are matrices of the probabilistic segmentation and the final segmentation maps respectively, $S$ is a symmetric normalized Laplacian matrix, $I$ is an identity matrix, and $\alpha$ is a parameter. Particularly, $S$ plays an important role in the label propagation in that it characterizes similarity among different voxels. In our study, image similarity between two voxels $x$ and $y$ is defined by

$$W_{xy} = \begin{cases} \exp\left(-\frac{\|I_x - I_y\|^2}{\sigma^2}\right), x \neq y \\ 0, x = y \end{cases}, \quad (3)$$

where $I_x$ and $I_y$ are image intensity values of voxels $x$ and $y$ respectively, and $\sigma$ is a parameter. Given a pairwise similarity matrix, $W$, between all voxels defined by Eq. (3),

$$S = D^{-1/2} W D^{-1/2}, \quad (4)$$

where $D$ is a diagonal matrix with its diagonal element equal to the sum of the corresponding row of $W$.

The optimization problem of Eq. (2) can be solved iteratively [21]:

$$L^{n+1} = (1 - \beta)SL^n + \beta L_0, \quad (5)$$

where $n$ is the number of iteration steps, and $0 < \beta < 1$ is a trade-off parameter related to $\alpha$.

To relieve impact of unreliable image segmentation results on the label propagation, the probabilistic segmentation map is weighted by reliable segmentation information, referred to as information-balance-weighting. Particularly, the reliable segmentation result is determined by a threshold ($|P| > T$). For segmenting the hippocampus, the number of background voxels is typical larger than the number of the hippocampal voxels. To balance the background and foreground label information in the label propagation, the reliable background label information is updated as

$$P_b = -max\left(\frac{N_f}{N_b}|P_b|, T\right), \quad (6)$$

where $N_f$ is the number of voxels with reliable foreground (hippocampus) segmentation labels, $N_b$ is the number of voxels with reliable background segmentation labels, $T$ is a threshold for determining the reliable segmentation result, and $P_b$ is a negative reliable segmentation label, i.e., $|P_b| > T$. It is worth noting that $max(\cdot, T)$ guarantees that the reliable $P_b$ remains greater than or equal to the threshold $T$ after the information-balance-weighting. Then, the probabilistic segmentation map is normalized separately for voxels with foreground and background probabilistic segmentation labels by $P_f = P_f / abs(\overline{P_f})$, and $P_b = P_b / abs(\overline{P_b})$, where $P_f$ is a positive segmentation label, $P_b$ is a negative segmentation label, and $\overline{P_f}$ and $\overline{P_b}$ are mean values of positive and negative labels respectively.

Finally, the weighted probabilistic segmentation map is updated using the label propagation method, as formulated by Eq. (5).

## 3. EXPERIMENTAL RESULTS

We have validated our method for the hippocampus segmentation based on imaging data and segmentation labels provided by the EADC–ADNI [23]. The dataset consists of a preliminary release part with 100 subjects and a final release part with 35 subjects. The final release data were used as training data and the preliminary release data were used as testing data. To segment the testing data, the images of training data were used atlases.

To reduce the computational cost, bounding boxes for both the left and right hippocampi were identified after all the images were linearly aligned to the MNI152 template with voxel size of $1 \times 1 \times 1$ mm$^3$. Instead of using all available atlases to segment a target image, 20 most similar atlases were selected based on normalized mutual information of the image intensities within the bounding box [11]. These atlases were then registered to the target image using ANTs [24]. To further reduce the computation cost, the majority voting label fusion method was used to obtain an initial segmentation result of the target image. Our method was then applied to voxels without 100% votes for either the hippocampus or the background in the majority voting method [11].

### 3.1 Parameter selection based on the training data

The image patches with size of $(7 \times 7 \times 7)$ were extracted from a neighborhood of $(3 \times 3 \times 3)$ following the LLL method [11]. Texture features were computed to build regression models using the RF regression method.

Our method has 6 parameters, including $k$ (the number of nearest neighboring samples for training local random forest), $N_{Tree}$ (the number of trees), $N_{Split}$ (the number of predictors sampled for splitting at each node), $T$ (the threshold for determining the reliable segmentation), $\sigma$ (the image similarity parameter), and $\beta$ (the trade-off parameter for updating the segmentation). Based on the training dataset, we adopted leave-one-out (LOO) cross-validation to optimize Dice index by grid-searching parameters from $k \in$ {100,200}, $N_{Tree} \in \{100, 200\}$, $N_{Split} \in \{10, 20, 30\}$, $T \in$ {0.3,0.4,0.5,0.6}, $\sigma \in \{5, 10, 20, 30\}$, and $\beta \in \{0.5, 0.6, 0.7, 0.8\}$. The optimal parameters were $k = 100$, $N_{Tree} = 200$, $N_{Split} = 20$, $T = 0.5$, $\sigma = 10$, and $\beta = 0.6$.

Since the MV method could provide probabilistic segmentation results too, we integrated the same SSLP method with the MV method for the hippocampus segmentation, referred to as MV-SSLP. The same parameters used in our method were adopted in the MV-SSLP method.

Table 1 summarizes segmentation performance of the MV, MV-SSLP, LLL-RF-regression, and RF-SSLP, measured by Dice index estimated using a LOO cross-validation based on the training dataset. The results demonstrated that the SSLP method could improve segmentation results of both the MV and the RF label fusion methods, and the integration of RF and SSLP obtained the best performance.

**Table 1.** Dice index values (mean±std) for 35 training subjects.

|  | Left hippocampus | Right hippocampus |
|---|---|---|
| MV | 0.859±0.03 | 0.863±0.02 |
| MV-SSLP | 0.867±0.03 | 0.871±0.02 |
| LLL-RF-regression | 0.885±0.02 | 0.886±0.01 |
| RF-SSLP | **0.890±0.02** | **0.891±0.01** |

### 3.2 Comparison with alternative MAIS methods

We compared our method with alternative MAIS methods, including majority voting (MV) [15], nonlocal patch (NLP) [16], local label learning (LLL) [11], random local binary pattern (RLBP) [8], and metric learning (ML) [7]. Parameters of these methods were set to optimal values suggested in their respective studies. Particularly, for the NLP method [16], its parameters were adaptively set using its suggested method; for the RLBP method [8], the number of generated RLBP features was set to 1000 and its balance parameter was set to $4^{-4}$; for the ML method [7], the number of nearest training samples was set to 9.

For the LLL method [11], we built both SVM classifiers and random forest classifiers, referred to as LLL-SVM and LLL-RF respectively. The image patches with size of $(7 \times 7 \times 7)$ were extracted from a neighborhood of $(3 \times 3 \times 3)$, and texture features were computed to build local classifiers based on 400 training samples. Sparse linear SVM classifiers were built with the default parameters of LibSVM, while random forest classifiers were built with following parameters: $k = 200$, $N_{Tree} = 200$, $N_{Split} = 20$, which were obtained to optimize Dice index by selecting parameters from $k \in \{100, 200\}$, $N_{Tree} \in \{100, 200\}$, and $N_{Split} \in \{10, 20, 30\}$.

Table 2 summarizes segmentation performance, measured by Dice index and mean distance (MD), of all the methods under comparison on both the training and testing datasets. For the training dataset, the segmentation performance was estimated based on a LOO cross-validation. Both the Dice index and measure distance measures indicated that the proposed method achieved superior segmentation performance relative to other methods under comparison.

Table 2. Dice index and Mean Distance (MD) values (mean±std) for 35 training subjects and 100 testing subjects obtained by different methods.

|      |          | Training data |           | Testing data |           |
|------|----------|---------------|-----------|--------------|-----------|
|      |          | left          | Right     | left         | right     |
| Dice | MV       | 0.859±0.03    | 0.863±0.02| 0.855±0.02   | 0.856±0.03|
|      | NLP      | 0.878±0.02    | 0.880±0.02| 0.870±0.02   | 0.873±0.02|
|      | LLL-SVM  | 0.882±0.02    | 0.884±0.02| 0.874±0.02   | 0.877±0.02|
|      | LLL-RF   | 0.885±0.02    | 0.886±0.02| 0.876±0.02   | 0.878±0.02|
|      | RLBP     | 0.884±0.02    | 0.886±0.02| 0.876±0.02   | 0.877±0.02|
|      | ML       | 0.884±0.02    | 0.886±0.02| 0.876±0.02   | 0.877±0.02|
|      | Proposed | **0.890±0.02**| **0.891±0.02**| **0.880±0.02**| **0.881±0.02**|
| MD   | MV       | 0.300±0.07    | 0.303±0.07| 0.303±0.05   | 0.320±0.07|
|      | NLP      | 0.255±0.04    | 0.265±0.05| 0.270±0.04   | 0.278±0.05|
|      | LLL-SVM  | 0.253±0.05    | 0.269±0.05| 0.263±0.03   | 0.279±0.05|
|      | LLL-RF   | 0.257±0.07    | 0.263±0.05| 0.263±0.03   | 0.279±0.05|
|      | RLBP     | 0.267±0.07    | 0.275±0.06| 0.270±0.04   | 0.290±0.06|
|      | ML       | 0.267±0.08    | 0.271±0.06| 0.271±0.05   | 0.287±0.06|
|      | Proposed | **0.240±0.04**| **0.251±0.05**| **0.254±0.03**| **0.270±0.05**|

## 4. CONCLUSIONS

We proposed a novel hippocampus segmentation method by integrating the random forest regression-based multi-atlas segmentation and the semi-supervised label propagation. Experiment results on the EADC-ADNI dataset demonstrated that our method achieved promising segmentation performance. Source codes of the algorithms under comparison are available at www.nitrc.org.

## ACKNOWLEDGEMENTS

This work was supported in part by the National Key Basic Research and Development Program of China [2015CB856404]; the National Natural Science Foundation of China [61473296]; the Promotive Research Fund for Excellent Young and Middle-Aged Scientists of Shandong Province [BS2014DX012]; China Postdoctoral Science Foundation [2015M581203]; the International Postdoctoral Exchange Fellowship Program [20160032]; and National Institutes of Health grants [EB022573, MH107703, DA039215, and DA039002].